\documentclass[]{article}
\usepackage{proceed2e}
\usepackage{amsmath, amssymb, graphicx}
\usepackage{natbib}

\title{Rank/Norm Regularization with Closed-Form Solutions: \\ Application to Subspace Clustering} 

\author{ {\bf Yao-Liang Yu}\\  
\texttt{yaoliang@cs.ualberta.ca}\\
Department of Computing Science\\
University of Alberta\\ 
Edmonton, AB, Canada, T6G 2E8
\And 
{\bf Dale Schuurmans} \\  
\texttt{dale@cs.ualberta.ca}\\
Department of Computing Science\\
University of Alberta\\ 
Edmonton, AB, Canada, T6G 2E8
}

\newcommand{\ui}[1]{\ensuremath{\|#1\|_{{\scriptscriptstyle \mathtt{UI}}}}} 
\newcommand{\uibig}[1]{\ensuremath{\left\|#1\right\|_{{\scriptscriptstyle \mathtt{UI}}}}} 
\newcommand{\aui}[1]{\ensuremath{\|#1\|_{{\scriptscriptstyle \mathtt{AUI}}}}} 
\newcommand{\auibig}[1]{\ensuremath{\left\|#1\right\|_{{\scriptscriptstyle \mathtt{AUI}}}}} 
\newcommand{\rui}[1]{\ensuremath{\|#1\|_{{\scriptscriptstyle \mathtt{RUI}}}}} 
\newcommand{\ruip}[1]{\ensuremath{\|#1\|_{{\scriptscriptstyle \mathtt{RUI'}}}}} 
\newcommand{\reg}[1]{\ensuremath{\|#1\|_{{\scriptscriptstyle \mathtt{REG}}}}} 
\newcommand{\uip}[1]{\ensuremath{\|#1\|_{{\scriptscriptstyle \mathtt{UI'}}}}} 
\newcommand{\uipbig}[1]{\ensuremath{\left\|#1\right\|_{{\scriptscriptstyle \mathtt{UI'}}}}} 
\newcommand{\fro}[1]{\ensuremath{\|#1\|_{{\scriptscriptstyle \mathtt{F}}}}} 
\newcommand{\spn}[1]{\ensuremath{\|#1\|_{{\scriptscriptstyle \mathtt{sp}}}}} 
\newcommand{\trn}[1]{\ensuremath{\|#1\|_{{\scriptscriptstyle \mathtt{tr}}}}} 
 
\newcommand{\rank}{\mathrm{rank}}
\newcommand{\diag}{\mathrm{diag}}

\newcommand{\st}{\mathrm{s.t.}}
\newcommand{\eproof}{$\mbox{}\null\hfill\blacksquare$}
\newtheorem{lemma}{Lemma}
\newtheorem{theorem}{Theorem}
\newtheorem{remark}{Remark}
\newtheorem{example}{Example}

\newtheorem{proposition}{Proposition}
\newtheorem{corollary}{Corollary}

\newcounter{thm}
\newtheorem{theoremappx}[thm]{Theorem}

\begin{document} 
 
\maketitle 
 
\begin{abstract} 
When data is sampled from an unknown subspace, principal component 
analysis (PCA) provides an effective way to estimate the subspace 
and hence reduce the dimension of the data.
At the heart of PCA is the Eckart-Young-Mirsky theorem, which 
characterizes the best rank $k$ approximation of a matrix. 
In this paper, we prove a generalization of the Eckart-Young-Mirsky theorem 
under all unitarily invariant norms.
Using this result, we obtain closed-form solutions for 
a set of rank/norm regularized problems,
%then 
and
derive 
%These results lead to 
closed-form solutions for a general class of subspace clustering problems
(where data is modelled by unions of unknown subspaces). 
From these results we obtain new theoretical insights 
and promising experimental results.

%Rank regularization is now understood to be an effective way to control model complexity. Since minimizing the rank in general is NP-Hard, in practice, one usually replaces the rank function by its convex envelop (restricted to the unit spectral norm ball), the trace norm. We show that a particular rank regularized problem, arising in the field of subspace clustering, admits a closed-form solution. Moreover, similar (trace) norm regularized problems are shown to have closed-form solutions as well. Our result significantly improves \cite{LRRICML, LRRPAMI} on subspace clustering and is expected to have applications in other problems, such as matrix completion.
\end{abstract} 
 
\section{Introduction} 
%Many real life data, 
Real world data,
while 
typically
being very high dimensional, %intrinsically only depend 
often only depends intrinsically 
on a few parameters \citep{Seung00}. It is %thus %very 
therefore
desirable to identify the low dimensional subspace (manifold) %where the data is sampled from. 
from which the data is sampled.
%The past a few years have witnessed an explosive efforts in that direction. 
Principal component analysis (PCA) \citep{PCA02} is a %classic, 
classical,
yet still popular, method to perform such dimensionality reduction. If the data is sampled from a \emph{single} subspace, PCA is provably correct in identifying the underlying subspace. Algorithmically, the success of PCA depends critically on the Eckart-Young-Mirsky theorem \citep{EckartYoung36, Mirsky60}, which characterizes, in %a 
closed-form, the optimal rank $k$ approximation of an arbitrary matrix, under all unitarily invariant norms.

However, in practice it is more likely that data is sampled, not from %one 
a \emph{single} subspace, but from a union of subspaces \citep{SC11}. Had we known the membership of the data, the task %is 
would be easy: we would just apply PCA to each subset. %subspace. 
%Unfortunately, more often than not, we do not have such useful membership information \emph{a priori}. % such precious membership information. 
Unfortunately, one does not normally have such useful membership information
\emph{a priori}.
%Thus 
The subspace clustering problem,
therefore,
is to find 
%is that of finding
the dimension and basis for each subspace %, %and segment/cluster 
while segmenting/clustering
the points accordingly. Of course, the %main 
primary
difficulty is that estimation and segmentation %need to be done 
must occur
\emph{simultaneously}, even though %any of the two tasks 
either task
can be easily accomplished given the result of the other. For %potential 
applications of %the 
subspace clustering, % problem, 
we refer the reader to the recent survey \citep{SC11}.

Many algorithms have been proposed for %the 
subspace clustering, % problem, 
including factorization methods \citep{Fact98, Fact06}, generalized principal component analysis (GPCA) \citep{GPCA05}, and agglomerative lossy compression \citep{ALC07}, as well as the more recent sparse subspace clustering (SSC) \citep{SSC09} and low rank representation (LRR) \citep{LRRICML}, %just 
to name a few. GPCA is provably correct while SSC and LRR are provably correct under the independent subspace assumption%, provided that the data is clean
. However, most current %existing 
algorithms are computationally extensive, requiring sophisticated numerical optimization routines. 

To develop simple algorithms for subspace clustering, we start by %from 
generalizing the %celebrated 
Eckart-Young-Mirsky theorem \citep{EckartYoung36, Mirsky60}, 
%as the latter is the building block of the elegant 
since it is the %primary basis 
foundation
for %the
PCA %algorithm 
in the single subspace scenario. 
In Section~2,
we prove a generalized version of the Eckart-Young-Mirsky theorem 
%in section 2, 
under all unitarily invariant norms. 
%Our 
The
motivation %to consider 
for considering
all unitarily invariant norms %comes 
%not only from 
is not solely for
mathematical completeness, it also arises
from the increasing popularity of the trace norm, %another 
%which is
a
special unitarily invariant norm. Using similar techniques, we are 
also able to provide closed-form solutions for some interesting rank/norm regularized problems. 
%Connection 
The subsequent connections
to known results are discussed.

In Section~3, we then apply the results %that have been 
established in Section~2 to the subspace clustering problem. Previous work %in 
by 
\citet{LRRICML} has proven that minimizing the trace norm of the reconstruction matrix yields a suitable block sparse matrix that will reveal the membership of the points in the ideal noiseless setting. We prove that this is not only true for the trace norm, but for essentially all unitarily invariant norms and the rank function. 
%Passing 
Progressing
to the noisy case, we propose to choose suitable combinations of unitarily invariant norms and the rank function in the objective, as it will lead to very simple algorithms that %which 
remain  provably correct if the data is clean and obeys the independent subspace assumption. Interestingly, the %proposed 
algorithms 
we propose
are intimately related to classic factorization methods \citep{Fact98, Fact06}. Experiments on both synthetic data and the hopkins155 motion segmentation dataset \citep{BenSeg07} %exhibit 
demonstrate
that the %proposed 
simple algorithms 
we devise
perform comparably with the state-of-the-art algorithms in subspace clustering, while being computationally much cheaper.

\textbf{Notations:} We use $\mathbb{M}^{m\times n}$ to denote $m\times n$ 
matrices, %though most times 
although generally 
we will not be very specific about the sizes. % of them. 
$\fro{\cdot}, \spn{\cdot}, \trn{\cdot}$ denote the Frobenius norm, spectral norm, and trace norm, respectively. For any matrix $A$, $A^*$ denotes its conjugate transpose, $A^\dag$ its pseudo-inverse, and $A_{(k)}$ its truncation, 
%by keeping 
where
the singular vectors 
are kept but
%and zeroing out 
all singular values 
are zeroed out
except the %first 
$k$ largest. % ones. 
%For a vector $a\in\mathbb{C}^{n}$, its $\ell_p$ norm is defined as $\|a\|_p = (\sum_{i=1}^n |a_i|^p)^{1/p}.$ We use $I$ to denote the identity matrix with suitable sizes.

%Due to space limitations, all proofs in the next section are omitted. Interested readers may consult a complete version downloadable from the authors' websites.
In this extended version of the paper, all proofs for the main results
stated in the next section are provided in the appendix.

\section{Main Results}
%We need some definitions first.  
First we require some definitions.
A matrix norm $\|\cdot\|$ is called unitarily invariant if $\|UAV\| = \|A\|$
%\forall 
for all
$A\in\mathbb{M}^{m\times n}$
%\forall \mbox{unitary matrices } 
and all unitary matrices
$U\in\mathbb{M}^{m\times m}$, 
$V\in\mathbb{M}^{n\times n}$. We use $\ui{\cdot}$ to denote unitarily invariant norms while $\aui{\cdot}$ means (simultaneously) $\emph{all}$ unitarily invariant norms.

Perhaps the most important examples for unitarily invariant norms are:
\begin{equation}
\label{uinorm}
\|A\|_{(k,p)} := \Big(\sum_{i=1}^k \sigma_i^p\Big)^{1/p},
\end{equation}
where $p\geq1$, $k$ is any natural number smaller than $\rank(A)$, and $\sigma_i$ is the $i$-th largest singular value of $A$. 
 For $k =\rank(A)$, (\ref{uinorm}) is known as Schatten's $p$-norm;  while
for $p = 1$, it is called Ky Fan's norm. Some special cases include the spectral norm ($p=\infty$), the trace norm ($p=1$, $k=\rank(A)$), and the Frobenius norm ($p=2$, $k=\rank(A)$).
Note that all three norms belong to the Schatten's family while only the first two norms are in the Ky Fan family. 

%The next two norms will also be useful for us:
%\begin{equation}
%\label{pqnorm}
%\|A\|_{p,q} := \Big(\sum_{j=1}^n \big(\sum_{i=1}^m |A_{ij}|^p\big)^{q/p}\Big)^{1/q},
%\end{equation}
%\begin{equation}
%\label{maxnorm}
%\maxn{A} := \min_{B^*C=A} \|B\|_{2,\infty}\cdot \|C\|_{2,\infty}.
%\end{equation}
%The $\|\cdot\|_{p,q}$ norm first takes the $\ell_p$ norm of each column, after which another $\ell_q$ norm is applied. (\ref{maxnorm}) is called the max norm since for positive semi-definite matrices, it reduces to the maximum of the diagonal. The max norm is known to be efficiently computable and has been used as another useful relaxation of the rank function, aside from the trace norm. We remind the reader for the subtle notational difference between (\ref{uinorm}) and (\ref{pqnorm}). Note that $\|\cdot\|_{2,q}$ is a left unitarily invariant norm while for any other values of $p$ and $q$, it is not left/right invariant. The max norm is not left/right invariant either.

The following theorem is well-known:
\begin{theorem}
\label{tEY}
Fix $\mathbb{N}\ni k\leq\rank(A)$, then 
$ A_{(k)}$ is a minimum Frobenius norm solution of 
\begin{equation}
\label{eEY}
\min_{X:~\rank{X}\leq k} \aui{A-X}.
\end{equation}
%It is the unique solution 
The solution is unique
iff the $k$-th and $(k+1)$-th largest singular values of $A$ differ.
\end{theorem}
Theorem \ref{tEY} was first%
\footnote{
%Note however that 
Erhard Schmidt proved a continuous analogue as early as 1907.
} 
proved by \citet{EckartYoung36} under the Frobenius norm; 
and then generalized to all unitarily invariant norms by \citet{Mirsky60}. The remarkable aspect %part 
of Theorem 1 is that although the rank constraint is highly nonlinear and nonconvex, %we are 
one is
still able to solve (\ref{eEY}) globally and efficiently by singular value decomposition (SVD). Moreover, the optimal solution under the Frobenius norm remains optimal under all unitarily invariant norms.

The Frobenius norm seems to be very different from other unitarily invariant norms, since % for 
it is induced by an inner product and block decomposable. Therefore it is usually much easier to work with the Frobenius norm, and much stronger results can be obtained
in this case. 
For instance, we have the following generalization of Theorem \ref{tEY} (though less well-known).
%, however, we should emphasize that, in terms of Theorem \ref{tEY}, any Schatten's $p$-norm, with $2 \leq p < \infty$, could replace the role played by the Frobenius norm, see Remark \ref{rem0}. 

For an arbitrary matrix $B$ with rank $r$, we denote its thin SVD as:
$B = U_B\Sigma_BV_{B}^*$. Define two projections 
$P_{B, \mathcal{L}} := U_{B} U_{B}^*$ and $P_{B, \mathcal{R}} := V_{B} V_{B}^*$.
Let $U_B^\perp$ and $V_B^\perp$ be the orthogonal complement of $U_B$ and $V_B$, respectively.
\begin{theorem}
\label{tGEY} 
Fix $\mathbb{N}\ni k\leq\rank(P_{B, \mathcal{L}} AP_{C, \mathcal{R}})$, then $B^{\dag} (P_{B, \mathcal{L}} AP_{C, \mathcal{R}})_{(k)} C^{\dag}$ is a minimum Frobenius norm solution of 
\begin{equation}
\label{eGEY}
\min_{X:~\rank{X}\leq k} \fro{A-BXC}.
\end{equation}
%It is the unique solution 
The solution is unique
iff the $k$-th and $(k+1)$-th largest singular values of $P_{B, \mathcal{L}} AP_{C, \mathcal{R}}$ differ.
\end{theorem}
Theorem \ref{tGEY} was first proved by \citet{GEY86}, but largely remained unnoticed. It was rediscovered recently by \citet{GEY07}. One may prove Theorem~\ref{tGEY} fairly easily, for instance, by adapting our proof for Theorem \ref{tLEY} below (plus the observation that the Frobenius norm is block decomposable). 

One natural question is whether we can replace the Frobenius norm in Theorem \ref{tGEY} with other unitarily invariant norms, as in Theorem \ref{tEY}. Unfortunately, Example 2 below shows that it is impossible in general. % impossible. 
However, by putting assumptions on $A, B$ and $C$, we are able to generalize Theorem \ref{tGEY} in meaningful ways.
%By putting assumptions on $A, B,$ and $C$, we are able to generalize Theorem \ref{tGEY} in a limited way (but strong enough for our application in low rank subspace clustering discussed in the next section).

\textbf{Simultaneous Block Assumption (SB):} Assume 
$(U_B^\perp)^* A V_C = 0$ and $U_B^* A V_C^\perp = 0$.
\begin{theorem} 
\label{tLEY}
Fix $\mathbb{N}\ni k\leq\rank(P_{B, \mathcal{L}} AP_{C, \mathcal{R}})$. 
Under the SB assumption, $B^{\dag} (P_{B, \mathcal{L}} AP_{C, \mathcal{R}})_{(k)} C^{\dag}$ is a minimum Frobenius norm solution of 
\begin{equation}
\label{eLEY}
\min_{X: ~\rank{X}\leq k} \aui{A-BXC}.
\end{equation}
%It is the unique solution 
The solution is unique
iff the $k$-th and $(k+1)$-th largest singular values of $P_{B, \mathcal{L}} AP_{C, \mathcal{R}}$ differ.
\end{theorem}
%Theorem \ref{tLEY} is essentially restricting $A$ to be in the range of $B$, $C$ to identity in Theorem \ref{tGEY}, but the norm has been successfully generalized to all unitarily invariant norms.
The next proposition plays a key role in the proof of Theorem \ref{tLEY}, and may be of some independent interest. %Both proofs can be found in the supplement.
\begin{proposition}
\label{prop}
If it exists, any minimizer of 
\begin{equation}
\label{con}
\min_{X\in\mathcal{X}} \aui{X}
\end{equation}
remains optimal for
$$\min_{X\in\mathcal{X}} 
\left\|\begin{pmatrix} X&0\\0 &B\end{pmatrix}\right\|_{{\scriptscriptstyle \mathtt{AUI}}}
$$
for any constant matrix $B$.
\end{proposition}
%\textbf{Proof:} see the supplement. \eproof

\begin{remark}
\label{rem1}
It is our incapability of %to extend 
extending
Proposition \ref{prop} to full block matrices, %matrix 
$\begin{pmatrix} X & C \\ D & B\end{pmatrix}$, 
that prevents us from fully generalizing Theorem \ref{tGEY}.
\end{remark}

One interesting case where the SB assumption is trivially satisfied %is 
can be
summarized as:
\begin{corollary}
\label{tCEY}
Fix $\mathbb{N}\ni k\leq\rank(A)$, then 
$B^{\dag} (BAC)_{(k)} C^{\dag}$ is a minimum Frobenius norm solution of 
\begin{equation}
\label{eCEY}
\min_{X: ~\rank{X}\leq k} \aui{BAC-BXC}.
\end{equation}
%It is the unique solution 
The solution is unique
iff the $k$-th and $(k+1)$-th largest singular values of $BAC$ differ.
\end{corollary}
%Putting 
Setting
$B$ and $C$ to identities, we recover Theorem~\ref{tEY}.
%We note 
Note
that a special case of %our 
this
corollary (where $B$ is identity and $C$ is a projection) has been 
previously 
established in \citep{RRE08, RRE09} in the context of reduced-rank estimators. However, our corollary is %considerably 
stronger %, and with 
and obtained by
a much simpler proof. 
%Our results might also lead to new insights on reduced-rank estimators, but we shall not pursue that direction any further in the current paper. 
We will apply Corollary \ref{tCEY} to subspace clustering in the next section.

%It is time to illustrate some examples.
We briefly illustrate %the consequences of 
these results with some examples.
\begin{example}
Consider $$A = \begin{pmatrix} 1 & 0 \\ 0 & 1\end{pmatrix},\; B = \begin{pmatrix} 1 \\ 0 \end{pmatrix},\; C = \begin{pmatrix} 1 & 0 \end{pmatrix}.$$
The SB assumption is satisfied, therefore one can %we may 
apply Theorem \ref{tLEY} to conclude that $x=1$ is the unique (rank 1) optimal solution under all unitarily invariant norms. However, %notice that 
Corollary \ref{tCEY} does not apply to this trivial example.
\end{example}

By now one might be tempted to hope that the SB assumption is just a %some 
removable artifact of the %our 
proof. This is not true: as the next example shows, the optimal solution under the Frobenius norm need not remain optimal under other unitarily invariant norms. This observation is in sharp contrast with Theorem \ref{tEY}.
\begin{example}
Consider $$A = \begin{pmatrix} 1 & 1 \\ 1 & 2\end{pmatrix},\; B = \begin{pmatrix} 1 \\ 0 \end{pmatrix},\; C = \begin{pmatrix} 1 & 0 \end{pmatrix}.$$ %Of course 
Here the SB assumption is  deliberately falsified. $X$ is now just a scalar and we require $\rank(X) \leq 1$. Under the Frobenius norm, it is easy to see $X=1$ is the (unique) optimal solution. However, under the trace norm, $X=0.5$ yields the optimal objective value 2.5, strictly better than $X=1$ whose objective value is $2\sqrt{2}$. Note that this example also demonstrates that Penrose's result, that is, $B^\dag A C^\dag$ is the minimum Frobenius norm solution of $\min_X \fro{A-BXC}$, does not generalize to other unitarily invariant norms (but see %compare with 
Remark \ref{rem4} below).
\end{example}

We currently do not know if problem (\ref{eGEY}), without the SB assumption, can or cannot be solved in polynomial time %once we replace 
if
the Frobenius norm 
is replaced
by any other unitarily invariant norm.%
\footnote{
We have found a positive result for the spectral norm. 
Details can be found in the complete version of %this 
the
paper.
} 

The last example shows that even one of $B$ and $C$ is restricted to identity, Theorem \ref{tGEY} still cannot be generalized to all unitarily invariant norms.
\begin{example}
Consider $$A = \begin{pmatrix} a & 0 \\ 0 & b \\ 0 & 1\end{pmatrix},\; B = \begin{pmatrix} 1 & 0 \\ 0 & 1 \\ 0 & 0\end{pmatrix},\; C = \begin{pmatrix} 1 & 0 \\ 0 & 1\end{pmatrix}.$$ Now $X$ is 2 by 2 and we require $\rank(X) \leq 1$. %Put 
Set
$a \vee b=1$ and $a\wedge b=1/2$. Under the Frobenius norm, it is easy to see $X=\begin{pmatrix} 1_{a>b} & 0 \\ 0 & 1_{a<b}\end{pmatrix}$ is the (unique) optimal solution. 
However, under the trace %(spectral, resp.) 
(resp.\ spectral) 
norm,
$X=\begin{pmatrix} a & 0 \\ 0 & 0\end{pmatrix}$ 
(resp.\ $X=\begin{pmatrix} 0 & 0 \\ 0 & b\end{pmatrix}$) %, resp.) 
yields 
a
strictly smaller objective value if $a<b$ 
(resp.\ $a>b$). %, resp.). 
Interestingly, %this time 
%here
Penrose's result, that %is, 
$B^\dag A$ is the minimum Frobenius norm solution of $\min_X \fro{A-BX}$, 
%{\em does\/} generalize 
generalizes
to all unitarily invariant norms 
in this case
\cite[Theorem 10.B.7]{Major11}.
\end{example}

%\begin{remark}
%\label{rem6}
%Instead of posing rank constraint on $X$, sometimes it might be useful to consider the following variant of (\ref{eLEY}):
%\begin{equation}
%\min_{X:~\rank{(BXC)}\leq k} \aui{A-BXC}.
%\end{equation}
%This problem can be easily reduced to (\ref{eLEY}) for if $k \geq \min\{\rank(B), \rank(C)\}$, the rank constraint is vacuous for $X$, which is the same as putting $k=\rank(A)$ in (\ref{eLEY}); otherwise it can be shown as equivalent to $\rank(\hat X) \leq k$ (see the proof of Theorem \ref{tLEY}), which eventually will lead to the same solution as (\ref{eLEY}).
%\end{remark}
%
%Another useful variant of (\ref{eLEY}) is when some columns of $A$ need to be fixed, a problem first considered by \cite{REY87}. Correspondingly, we have the following strict generalization of \cite{REY87}.
%\begin{theorem} 
%\label{tREY}
%Under the SB assumption, the minimum Frobenius norm solution of 
%\begin{equation}
%\label{eREY}
%\!\min_{X:~\rank{[A_1~ X]}\leq k} \aui{B[\begin{pmatrix}A_1 & A_2\end{pmatrix} - \begin{pmatrix} A_1 & X \end{pmatrix}]C}.\!
%\end{equation}
%is 
%$$X^\star = \left\{\begin{array}{cc}
%B^{\dag} (P_{B, \mathcal{L}} AP_{C, \mathcal{R}})_{(k)} C^{\dag} & k-\rank(A_1) \geq \min\{\rank B, \rank C\}\\ bla & \mbox{otherwise}
%\end{array}\right.$$
%\end{theorem}
%Notice that the subtlety of (\ref{eREY}) is that $\rank{[A_1~ X]}\leq k$ in general is not equivalent to $\rank(X) \leq k-\rank(A_1)$. Putting $B, C$ to identities we recover the main Theorem in \cite{REY87}.
%
%\textbf{Proof:}
%\eproof

\begin{remark}
\label{rem2} 
So far, we have restricted %ourselves to solve 
attention to %solving
rank constrained problems. Fortunately, rank {\em regularized\/} problems 

\vspace*{-2ex}

\begin{equation}
\label{rr}
\min_{X\in\mathbb{M}^{m\times n}} f(X) + \lambda\cdot\rank(X)
\end{equation}

\vspace*{-1ex}

can always be efficiently reduced to %its 
an equivalent
constrained version

\vspace*{-2ex}

\begin{equation}
\label{rc}
\min_{X:~\rank(X)\leq k} f(X), 
\end{equation}

\vspace*{-1ex}

since the rank function can only take integral values between 0 and $\min\{m,n\}$. 
%We 
Here
one
need only solve (\ref{rc}) for each admissible value of $k$, % and 
then pick the best according to the objective of (\ref{rr}). Hence, it is clear that if %we 
one
can efficiently solve (\ref{rc}) for all admissible values of $k$, 
then (\ref{rr}) can be efficiently solved
%we can efficiently solve (\ref{rr}) 
for all values of $\lambda$ (even negative ones, which promote the rank).
\end{remark}

%Also, 
Note that,
due to the discreteness of the rank function, 
%in tuning the constant $\lambda$, 
the optimal solution changes discontinuously
when tuning the constant $\lambda$.
%leading to instability. 
%hence 
Therefore,
it is usually desirable to \emph{smooth} the solution, even when one %we 
can optimally solve the rank regularized problem. This is usually done by replacing the rank function with a suitable norm. 
%Our 
The
next theorem %says 
states
that Theorem \ref{tLEY} has a close counterpart for unitarily invariant norms, although under %some 
a
stronger assumption.

\textbf{Simultaneous Diagonal Assumption (SD):} In addition to the SB assumption, assume furthermore 
that
$U_B^*AV_C$ is diagonal.%
\footnote{
Rectangular matrix $A$ is diagonal if $A_{ij} = 0,~\forall i\ne j$.
}
\begin{theorem}
\label{tNEY}
Let $\lambda > 0$. Under the SD assumption, the matrix problem
\begin{equation}
\label{eNEY}
\min_{X} \ui{A-BXC} + \lambda\cdot\uip{X} 
\end{equation}
has an optimal solution of the form $X^\star = V_B\Sigma_X U_C^*$, 
%with 
where
$\Sigma_X$ %being 
is
diagonal.
\end{theorem}
Note that the two unitarily invariant norms in (\ref{eNEY}) need not be the same. %The proof of the theorem can be found in the supplement.

\begin{remark}
\label{rem3}
A couple of observations are in order:
\begin{itemize}
\item The SD assumption is considerably stronger than the SB assumption. This is because the rank function has more invariance properties that we can exploit: it is not only unitarily invariant, but also scaling invariant. 
In contrast, norms, by their definition, cannot be scaling invariant.
\item Unlike the rank constrained problem (\ref{eLEY}), the norm regularized problem (\ref{eNEY}) can always be solved in polynomial time as long as %we 
one
can evaluate the norms in polynomial time. After all (\ref{eNEY}) is a convex program. However, the point of Theorem \ref{tNEY} is to characterize situations where %we can solve 
the problem 
can be solved
in nearly closed-form.
\item Sometimes rather than regularizing, one might prefer to constrain %ing 
the norm to be smaller than some constant. %We %may 
One can
easily adapt the proof of Theorem \ref{tNEY} %for its 
to the
constrained version. Moreover, by choosing suitable constants, regularized problems and their constrained counterparts can yield the same solutions. 
%We will only consider regularized problems in this paper for convenience.
\item Theorem \ref{tNEY} obviously remains true if %we put 
one asserts
(possibly different) monotonically increasing transforms %in front of 
around
the two norms.
\end{itemize}
\end{remark}

We now discuss two interesting cases where the SD assumption is trivially satisfied. Let $A = \sum_{i=1}^{\rank(A)} \sigma_i U_iV_i^*$ be %its 
the
thin SVD.
\begin{corollary}
\label{matcom}
Let $\lambda > 0$. The matrix problem 
$$\min_{X} \|A-X\|_{(k,p)} + \lambda\cdot\|X\|_{(k',p')}$$ has a (nearly) closed-form solution $X^\star = \sum_{i=1}^{\rank(A)} x_i^\star U_iV_i^*$, where $x^\star$ solves 
$$\min_{x\in\mathbb{R}_+^{\rank(A)}} \Big[\sum_{i=1}^k |\sigma - x|_{[i]}^p\Big]^{1/p} + \lambda\cdot \Big[\sum_{i=1}^{k'} x_{[i]}^{p'}\Big]^{1/p'}.$$
(Here $x_{[i]}$ %means 
is
the $i$-th largest element of the vector $x$.)
\end{corollary}
If we %put 
set
the first norm in Corollary \ref{matcom} to %be 
the squared Frobenius norm (see the last item in Remark \ref{rem3}) %, 
and the second %one 
%norm
to the trace norm, %we then recover 
%one then recovers
we recover
the SVD thresholding algorithm %in 
for
matrix completion \citep{SVT10,SVT09}. Note that the existing 
correctness
proof %of 
for
%the 
SVD thresholding %algorithm 
relies on the complete characterization of the subdifferential of the trace norm, while our proof takes a very different path and avoids deep results in convex analysis.%
\footnote{Upon completing %finishing this 
the paper, we %noted 
discovered
that a similar argument tailored for the trace norm 
has
appeared in \citep{LRRICDM}.
} 
One %may 
can also
easily derive closed-form solutions for other variants, for instance, if the first norm is the $p$-th power of Schatten's $p$-norm while the regularizer is still the trace norm, %we get 
similar thresholding algorithms ensue.
\begin{corollary} 
\label{subclu}
Let $\lambda > 0$. The matrix problem 
$$\min_{X} \|A-AX\|_{(k,p)} + \lambda\cdot\|X\|_{(k',p')}$$ has a (nearly) closed-form solution $X^\star = \sum_{i=1}^{\rank(A)} x_i^\star V_iV_i^*$, where $x^\star$ solves 
$$\min_{x\in\mathbb{R}_+^{\rank(A)}} \Big[\sum_{i=1}^k |\sigma - \sigma\odot x|_{[i]}^p\Big]^{1/p} + \lambda\cdot \Big[\sum_{i=1}^{k'} x_{[i]}^{p'}\Big]^{1/p'}.$$
%Again 
Here
$x_{[i]}$ denotes the $i$-th largest element and $\odot$ is the Hadamard (elementwise) product.
\end{corollary}
We %will 
apply Corollary \ref{subclu} to the subspace clustering problem in the next section.

We close this section with a reversed version of Corollary \ref{tCEY}. 
Let $B = U_B\Sigma_BV_B^*$ and $C=U_C\Sigma_CV_C^*$ be the %ir 
corresponding
thin SVDs. Define $\hat A:= V_B^*AU_C$.
\begin{theorem}
\label{tranknorm}
Let $\lambda > 0$, then $\exists~ r\in\{0,\ldots,\rank(\hat A)\}$ such that $V_B(\hat A - \hat A_{(r)})U_C^*$ is a minimum Frobenius norm solution of
\begin{equation}
\label{eranknorm}
\min_X \rank(BAC-BXC) + \lambda \rui{X},
\end{equation}
where $\rui{\cdot}$ is either the rank function or a unitarily invariant norm.
\end{theorem}
%The proof can be found in the supplement.

\begin{remark}
\label{rem4}
Theorem \ref{tranknorm} also implies a closed-form solution for the following problem:
\begin{equation}
\label{eqtmp}
\min_{X: A=BXC} \rui{X}.
\end{equation} 
By a classic result of \citet{Penrose56}, if the feasible set is not empty,%
\footnote{
Of course, this can be easily and efficiently checked.
}
then $B^\dag A C^\dag$ is a feasible point, hence we may write $A$ as $BB^\dag A C^\dag C$. 
%Use 
Using
$B^\dag A C^\dag$ as $A$ in (\ref{eranknorm}), and %let 
letting
$\lambda \to 0$, %we find 
one obtains
the closed-form solution for (\ref{eqtmp}): $X^\star = B^\dag AC^\dag$%, which is exactly Penrose's result (\cite{Penrose56}) if $\reg{\cdot} = \fro{\cdot}$
. Another way to derive this fact is through Corollary \ref{tCEY} by setting $k$ large. With slightly more %efforts 
effort
one can also prove the uniqueness of the solution if $\rui{\cdot}$ %belongs to  
is a
%Schatten's 
Schatten
$p$-norm ($p < \infty$). When $\rui{\cdot} = \fro{\cdot}$, we recover the classic result of \citep{Penrose56}; for $\rui{\cdot} = \trn{\cdot}$, we recover the recent result in \citep{LRRPAMI};%
\footnote{
Although the formula in Theorem 3.1 of \citep{LRRPAMI} looks different from ours, it can be verified that they are indeed the same.
}
and for $\rui{\cdot} = \rank(\cdot)$, we recover a (weaker) result in \citep{Tian03}. %To our surprise, 
Surprisingly,
all other cases appear to be new.
\end{remark}

\section{Subspace Clustering}

%Formally, 
Subspace clustering considers the following problem: 
Given a set of points $X = [X_1, \ldots, X_k]\Gamma$ in $\mathbb{R}^D$, where $X_i = [x^i_1, \ldots, x^i_{n_i}]$ is drawn from some unknown subspace $\mathcal{S}_i$ with unknown dimension $d_i$ 
(i.e., $x_j^i\in\mathbb{R}^D$ is the $j$-th sample from subspace $\mathcal{S}_i$) and $\Gamma$ is an unknown permutation matrix; we want to identify the number of subspaces $k$, the dimension $d_i$ and basis $V_i$ for each subspace, %and segment 
while simultaneously segmenting
the points accordingly (i.e., %estimate 
estimating
$\Gamma$ and $n_i$). In general, this is a ill-posed problem, but if some prior knowledge of $k$ (the number of subspaces) or $d_i$ (the dimension of subspace $\mathcal{S}_i$) is provided, %we may 
one can
solve the subspace clustering problem in a meaningful way. For example, if we assume $k=1$, then subspace clustering reduces to %the 
classic principal component analysis, which has been well-studied and widely applied.

Subspace clustering is very challenging %since we need 
because one has
to \emph{simultaneously} estimate the subspaces and segment the points, even though %any one of them 
each subproblem
can be easily solved given the result of the other. Practical issues like computational complexity, noise, and outliers make the problem even more challenging. We refer the reader to the excellent survey \citep{SC11} for details.

Recently, under the \emph{independent}%
\footnote{
A set of subspaces is called independent if the dimension of their direct sum equals the sum of their dimensions.
} 
subspace assumption (and %the data is clean), 
assuming clean data),
%\citep{SSC09} 
\citet{SSC09} 
successfully %built 
recover
a block sparse matrix to reveal %the 
data
membership, % of the data, 
by resorting to the sparsest reconstruction of each point %through 
from
other points. The key observation is that each point can only be represented by other points from the same subspace, due to the independence assumption. 
%\citep{LRRICML, LRRPAMI} 
\citet{LRRICML, LRRPAMI} 
subsequently showed that similar block sparse matrix can be obtained by minimizing the trace norm% of the reconstruction matrix
, instead of the $\ell_1$ norm in \citep{SSC09}.

Specifically, 
%\citep{LRRICML} 
\citet{LRRICML} 
considered the following problem:% first:%
\footnote{
Instead of the data $X$ itself, in principle one could also choose other dictionaries to reconstruct $X$. As long as the dictionary spans $X$, our results in this section still apply.
}
\begin{equation}
\label{ranksub}
\min_Z~ \rank(Z)   \quad\st\quad X = XZ.
\end{equation}
The idea %behind (\ref{ranksub}) 
is that, 
given the independence assumption, 
if %we 
one
reconstructs 
each point through other points, % due to 
%given the independence assumption, 
the reconstruction matrix $Z$ must have low rank. However, (\ref{ranksub}) was thought to be hard, hence 
%\citep{LRRICML} 
\citet{LRRICML} 
turned to %its 
a
convex relaxation:
\begin{equation}
\label{normsub}
\min_Z~ \trn{Z} \quad\st\quad X=XZ.
\end{equation}
Under the independence assumption, 
%\citep{LRRICML} 
\citet{LRRICML} 
successfully proved that $Z_{ij} = 0$ if points $x_i$ and $x_j$ come from different subspaces. Our result in Remark \ref{rem4} then immediately yields a generalization to all unitarily invariant norms (and the rank function, which was thought to be hard in \citep{LRRICML}). Recall that we use $\rui{\cdot}$ to denote either the rank function or an arbitrary unitarily invariant norm.
\begin{theorem}
\label{subth}
Assume the subspaces are independent and the data is clean, then $Z^\star:=X^\dag X$, being a minimum Frobenius norm solution of 
\begin{equation}
\label{regsub}
\min_Z~ \rui{Z} \quad\st\quad X=XZ,
\end{equation}
is block sparse, that is,  $Z^\star_{ij} = 0$ if points $x_i$ and $x_j$ come from different subspaces.
\end{theorem}
\textbf{Proof:} As shown in \citep{LRRICML}, (\ref{regsub}) under the trace norm has a unique solution that satisfies the block sparse property.  But Remark \ref{rem4} shows that $X^\dag X$ is %this 
the
unique solution and moreover %it 
remains optimal under all unitarily invariant norms and the rank function. 
\eproof

%\begin{remark}
%In fact, many more norms can yield Theorem \ref{subth}. For instance, all $\ell_p/\ell_q$ norm\footnote{The $\ell_p/\ell_q$ norm for matrix $A$ is defined as: $[\sum_j(\sum_i A_{ij}^p)^{q/p}]^{1/q}$.} ($p, q<\infty$). We believe it is the independence assumption, not the magic of trace norm, that leads to the nice result in Theorem \ref{subth}.
%\end{remark}
We noted that $X^\dag X$ is called the shape interaction matrix (SIM) in the computer vision literature, and was known to have the block sparse structure \citep{Fact98}. The surprising aspect 
of Theorem \ref{subth} is that, 
at least in the ideal noiseless case, there is nothing special about the trace norm. Any unitarily invariant norm, in particular, the Frobenius norm, leads to the same closed-form solution.

Of course, in practice, data is always corrupted by noise and possibly outliers. To account for %that, we may consider:
this, one can consider:
\begin{equation}
\label{noisy}
\min_Z ~ \rho(X-XZ) + \lambda\cdot\reg{Z},
\end{equation}
where $\rho(\cdot)$ measures the discrepancy between $X$ and $XZ$, $\reg{\cdot}$ is a regularizer, and $\lambda$ is the parameter that balances the two terms. 
In this case,
popular choices of $\rho$ include the (squared) Frobenius norm, $\ell_1$ norm, the $\ell_2/\ell_1$ norm or even the rank function, depending on our assumption of the noise. Typical regularizers include the trace norm, Frobenius norm or the rank function. For instance, \citet{LRRPAMI} considered $\rho = \ell_1$ (if the noise is sparsely generated) or $\rho = \ell_2/\ell_1$ (if the noise is sample specific), and $\reg{\cdot} = \trn{\cdot}$. %They solved 
The resulting convex program 
was solved
by the method of augmented Lagrangian multipliers.

When %we do not have much 
such
prior information about the noise
is not available, 
it becomes a matter of subjectivity to choose $\rho$ and $\reg{\cdot}$. Our next result shows that, by choosing $\rho$ and $\reg{\cdot}$ appropriately, %we may have 
one can still obtain
a closed-form solution for (\ref{noisy}):
\begin{theorem}
\label{DSSIM}
Let $X = U\Sigma V^*$ be %its 
the
thin SVD. %then 
Then
$\exists r \in\{0,\ldots,\rank(X)\}$ such that $V_{(r)}V_{(r)}^*$ is a minimum Frobenius norm solution of 
$$\min_Z \rui{X- XZ} + \lambda \cdot\ruip{Z},$$
where one of $\rui{\cdot}$ and $\ruip{\cdot}$ is the rank function, or both are the trace norm.
\end{theorem}
\textbf{Proof:} The case $\ruip{\cdot}=\rank$ follows from Corollary \ref{tCEY} and Remark \ref{rem2}; the case $\rui{\cdot} = \rank$ follows from Theorem \ref{tranknorm}; and the last case $\rui{\cdot}=\ruip{\cdot}=\trn{\cdot}$ follows from Corollary \ref{subclu}. \eproof

Interestingly, $V_{(r)}V_{(r)}^*$ was known to be %a good heuristic to handle 
an effective heuristic for handling
noise in the computer vision literature \citep{Fact06}. 
%However, here it is the first time to justify this 
Here we provide a %first 
formal justification for this
heuristic by showing that it is an optimal solution of some reasonable optimization problem(s). This new interpretation is important for model selection purposes in the unsupervised setting. Intuitively, the idea behind $V_{(r)}V_{(r)}^*$ is also simple: If the amount and magnitude of noise is moderate, the SIM will not change %much, 
significantly
hence by thresholding %out 
small singular values, which usually are caused by noise, %we 
one
might still be able to recover the SIM, approximately. We shall call $V_{(r)}V_{(r)}^*$ the discrete shrinkage shape interaction matrix (DSSIM).

%As we remarked before, 
As remarked previously,
the discrete %ness 
nature of the DSSIM might lead to instability, hence it might be %useful 
preferable
to consider the following variant
\begin{equation}
\min_Z ~\ui{X-XZ} + \lambda\cdot\uip{Z},
\end{equation} 
which has also been shown to have a (nearly) closed-form solution in Corollary \ref{subclu}. Specifically, we have the following result:
\begin{corollary}
Let $X=\sum_i \sigma_i U_iV_i^*$ be %its 
the
thin SVD. 
Then 
$\sum_i (1-\frac{\lambda}{2\sigma_i^2})_+V_iV_i^*$ is an optimal solution of 
\begin{equation}
\min_Z ~\fro{X-XZ}^2 + \lambda\cdot\trn{Z},
\end{equation} 
where $(\cdot)_+=\max(0,\cdot)$ %means 
denotes
the positive part.
\end{corollary}
We shall call the solution in the above corollary the continuous shrinkage shape interaction matrix (CSSIM). For comparison purposes, we also consider $\sum_i \frac{\sigma_i^2}{\sigma_i^2+\lambda}V_iV_i^*$, the closed-form solution of 
\begin{equation}
\min_Z ~\fro{X-XZ}^2 + \lambda\cdot\fro{Z}^2.
\end{equation} We shall call it the smoothed shape interaction matrix (SSIM).

%At last, 
Finally,
we show that for all choices of the discrepancy measure $\rho$, and regularizers $\rui{\cdot}$ of the rank function or a unitarily invariant norm, the optimal solutions of (\ref{noisy}) share some common structure. To see %that, 
this
let us consider the equivalent problem:
\begin{equation}
\min_{Z, R}~ \rho(R) + \lambda\cdot\rui{Z} \quad \st \quad X = XZ + R.
\end{equation}
The first observation is that $R$ must lie in the range space of $X$ due to the equality constraint, hence we %may 
can
let $R:=XE$. Moreover, given $R$, using results in Remark \ref{rem4}, we obtain $$Z=X^\dag (X-R) := X^\dag X(I-E).$$
Therefore, we see that no matter how we choose $\rho$, the resulting optimal solution is a modification of the SIM.

\section{Experiments}
In this section, we compare the closed-form solutions (SIM, DSSIM, CSSIM and SSIM) derived here with the low rank subspace clustering algorithm proposed in \citep{LRRICML,LRRPAMI}. 
%as the 
The
latter has been shown to achieve the state-of-the-art for subspace clustering problems. Specifically, two variants in \citep{LRRPAMI}, which we denote LRR1 ($\rho$ is the $\ell_2/\ell_1$ norm) and LRR2 ($\rho$ is the $\ell_1$ norm), respectively, are compared. For all methods%
\footnote{
For DSSIM, we use the objective in Theorem \ref{DSSIM} (the trace norm case) to tune $\lambda$ as we find this yields better performance than tuning %directly 
the rank $r$
directly.
} 
except SIM, we tune the regularization constant $\lambda$ within the range $\{10^i, i=-4:1:4\}$. SIM does not have such a parameter (i.e., $\lambda\equiv 0$).

After obtaining the reconstruction matrix $Z$, an affinity matrix $W_{ij} = |Z_{ij}| + |Z_{ji}|$ is built. Standard spectral clustering techniques \citep{Luxburg07} are applied to segment the points into different clusters (subspaces). We count the number of misclassified points, and report the accuracy of each method.

\subsection{Synthetic Data}
\begin{figure*}[t]
\centering
\includegraphics[width=1.5\columnwidth]{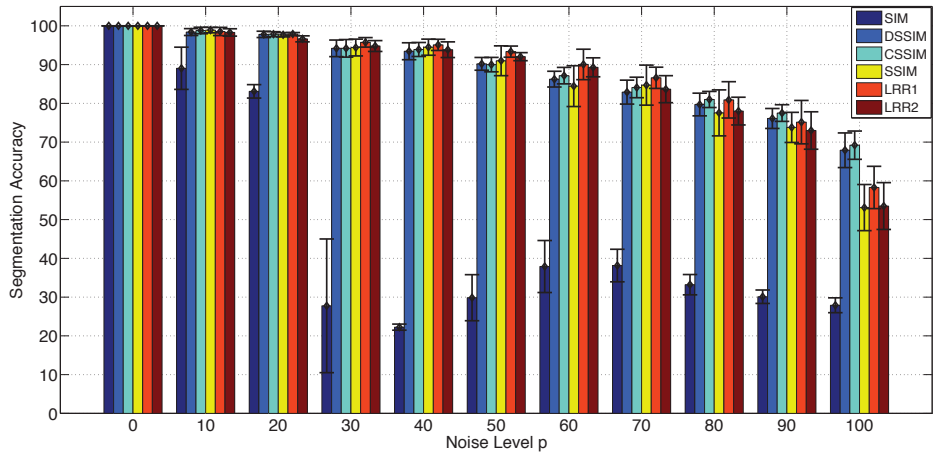}
\caption{Results for synthetic data.}
\label{fig}
\end{figure*}

Following \citep{LRRPAMI}, 5 independent random subspaces $\{\mathcal{S}_i\}_{i=1}^5 \subseteq \mathbb{R}^{100}$, each with dimension 10, are constructed. Then 40 points are randomly sampled from each subspace. We randomly choose $p\%$ points and corrupt them with zero mean Gaussian noise, whose standard deviation is 0.3 times the length of the point. 
%Note that the independent subspace assumption is satisfied in this synthetic dataset. 
We repeat the experiment 10 times and the averaged results are reported in Figure \ref{fig}.

When $p=0$, i.e., in the ideal noiseless setting, all methods achieve perfect segmentation, as expected from Theorem \ref{subth}. As we increase the noise level $p$, SIM degrades quickly since it has no protection against noise. LRR1 performs best in the range of $p=30\sim 70$, probably because its discrepancy measure matches the noise generation process the best. However, we note that the advantage of LRR1 over other methods is rather small. When most data points are corrupted ($p=80\sim 100$), DSSIM and CSSIM start to prevail. 
Overall, CSSIM, based on the trace norm regularizer, performs slightly better than SSIM, which is based on the Frobenius norm regularizer. 

On the computational side, SIM, DSSIM, CSSIM and SSIM all have closed-form solutions and %need only do SVD once, 
only require 
a single 
%one
call to
SVD,
while LRR generally requires 300 %more 
%additional
steps to converge on this dataset; 
that is, 300 SVDs, since each step involves the SVD thresholding algorithm. Moreover, LRR pays extra computational cost in selecting the regularization constant. In total, LRR is %at least 2,700 times 
orders of magnitude
slower than all other methods.
%than the SIM methods.

We noted in experiments that if the noise magnitude is larger than a threshold, all methods will degrade to the level of SIM.

\subsection{Hopkins155 Motion Segmentation}
The hopkins155 dataset is a standard benchmark for motion segmentation and subspace clustering \citep{BenSeg07}. It consists of 155 sequences of two or three motions. The motions in each sequence are regarded as subspaces while each sequence is regarded as a separate clustering problem, resulting in total 155 subspace clustering problems. The outliers in this dataset has been manually removed, hence we expect the independent subspace assumption to hold approximately. For 
a
fair comparison, %s, 
we apply all methods to the raw data, even though we have observed in experiments that the performances can be further improved by suitable preprocessing/postprocessing.

The results are tabulated in Table \ref{tab}. All methods perform well on this dataset, even SIM achieves 75.56\% accuracy, confirming that this dataset only contains small amount/magnitude of noise and obeys the independent subspace assumption reasonably well. The numbers in the ``Mean'' row are obtained by first averaging over all 155 sequences and then selecting the best out of the 9 regularization constants, with the best $\lambda$ tabulated below. The results are close to the state-of-the-art as reported in \citep{BenSeg07}. If we are allowed to select the best $\lambda$ individually for each sequence, we obtain the ``Best'' row, which is surprisingly good. 
It is clear that LRR is significantly slower than all other methods. Note that the running time is not averaged over sequences, nor does it include the spectral clustering step or the tuning step.
\begin{table}[t]
\vspace{-0.8em}
\centering
\scriptsize{
\caption{Segmentation Accuracy (\%) on Hopkins155.}
\label{tab}
\begin{tabular}{c|c|c|c|c|c|c}
\hline
Method & SIM & DSSIM & CSSIM & SSIM & LRR1 & LRR2\\
\hline
Mean & 75.56 & 95.51 & 96.05 & 96.82 & 96.37 & 96.52\\
$\lambda$ & 0 & $10^{-2}$ & $10^{-3}$ & $10^{-2}$ & 10 & 1\\
Time & 3.56s & 3.29s & 3.61s & 3.61s & 695.1s & 734.6s\\
\hline
Best & 75.56 & 99.27 & 99.29 & 99.46 & 99.40 & 99.32\\
\hline
\end{tabular}}
\vspace{-1.em}
\end{table}

\section{Conclusion}
We have generalized the celebrated Eckart-Young-Mirsky theorem, under all unitarily invariant norms. Similar techniques are used to provide closed-form solutions for some interesting rank/norm regularized problems. The results are applied to subspace clustering, resulting in very simple algorithms. Experimental results demonstrate that the proposed algorithms perform comparably against the state-of-the-art in subspace clustering, but with a significant computational advantage.

%\newpage
\small{\bibliographystyle{apalike}}
\bibliography{bib} 

\newpage
\appendix

{\bf\large Appendix: Proofs of the Main Results}

\section{Preliminaries}
%We recall some definitions first.  
We first re-establish some definitions from the main body.
A matrix norm $\|\cdot\|$ is called unitarily invariant if $\|UAV\| = \|A\|$
for all $A\in\mathbb{M}^{m\times n}$ and all unitary matrices 
$U\in\mathbb{M}^{m\times m}$ and $V\in\mathbb{M}^{n\times n}$. 
We use $\ui{\cdot}$ to denote unitarily invariant norms while 
$\aui{\cdot}$ means (simultaneously) $\emph{all}$ unitarily invariant norms.
The notation $:=$ is used to indicate a definition.

As mentioned, the most important examples for unitarily invariant norms are
perhaps:
\begin{equation}
\label{uinormappx}
\|A\|_{(k,p)} := \Big(\sum_{i=1}^k \sigma_i^p\Big)^{1/p},
\end{equation}
where $p\geq1$, $k$ any natural number smaller than $\rank(A)$, and $\sigma_i$ is the $i$-th largest singular value of $A$. 
 For $k =\rank(A)$, (\ref{uinormappx}) is known as Schatten's $p$-norm;  while
for $p = 1$, it is called Ky Fan's norm. Some special cases include the spectral norm ($p=\infty$), the trace norm ($p=1$, $k=\rank(A)$), and the Frobenius norm ($p=2$, $k=\rank(A)$).
Note that all three norms belong to the Schatten's family while only the first two norms are in the Ky Fan family. However, Ky Fan's norm turns out to be very important in studying general unitarily invariant norms, due to the following fact (Theorem V.2.2, \cite{Bhatia97}):
\begin{lemma}
\label{KyFan}
$\aui{A} \leq \aui{B}$ iff $\forall k, \|A\|_{(k,1)} \leq \|B\|_{(k,1)}$.
\end{lemma}
Another interesting fact about unitarily invariant norms is (Problem II.5.5, \cite{Bhatia97}):
\begin{lemma}
\label{diagdom}
$$\Bigg\|\begin{pmatrix} A & B \\ C & D \end{pmatrix}\Bigg\|_{\footnotesize \mathtt{UI}} \geq \Bigg\|\begin{pmatrix} A & 0 \\ 0 & D \end{pmatrix}\Bigg\|_{\footnotesize \mathtt{UI}}
\geq \Bigg\|\begin{pmatrix} A & 0 \\ 0 & 0 \end{pmatrix}\Bigg\|_{\footnotesize \mathtt{UI}}$$
\end{lemma}
Notice that \cite{Bhatia97} assumes $A$ and $D$ to be square matrices. This assumption  may be easily removed by padding with zeros. 
%It is also easy to show that the inequality is strict for Schatten's $p$-norm, $2 \leq p < \infty$, unless $B=0, C=0$. 
It is clear by induction that Lemma \ref{diagdom} can be extended to any number of blocks.

The following theorem is well-known and its proof can be found in \cite{Mirsky60}:
\begin{theoremappx}
\label{tEYappx}
Fix $\mathbb{N}\ni k\leq\rank(A)$, then 
$ A_{(k)}$ is the minimum Frobenius norm solution of 
\begin{equation}
\label{eEYappx}
\min_{X:~\rank{X}\leq k} \aui{A-X}.
\end{equation}
%It is the unique solution 
The solution is unique
iff the $k$-th and $(k+1)$-th largest singular values of $A$ differ.
\end{theoremappx}

\section{Proofs}
We first prove the key proposition.
\begin{proposition}
\label{propappx}
If it exists, any minimizer of 
\begin{equation}
\label{conappx}
\min_{X\in\mathcal{X}} \aui{X}
\end{equation}
remains optimal for
$$\min_{X\in\mathcal{X}} \auibig{\begin{pmatrix} X&0\\0 &B\end{pmatrix}}$$
for any constant matrix $B$.
\end{proposition}
\textbf{Proof:} 
The proof is to repeatedly apply Lemma \ref{KyFan}. 
Recall that the Ky Fan norm $\|\cdot\|_{(k,1)}$ defined in (\ref{uinormappx}) is the sum of the $k$ largest singular values. 
Let $X^\star$ be an optimal solution of (\ref{conappx}). 
According to Lemma \ref{KyFan}, $\|X^\star\|_{(k,1)} \leq \|X\|_{(k,1)}$ 
for all admissible values of $k$ and for all feasible $X\in\mathcal{X}$. 
Then for all admissible values of $k$ and for all feasible $X\in\mathcal{X}$
we have:
\begin{eqnarray*}
\left\|\begin{pmatrix} X&0\\0 &B\end{pmatrix}\right\|_{(k,1)} &:=& \|X\|_{(k_1,1)} + \|B\|_{(k_2,1)}\\
&\geq& \|X\|_{(\hat k_1,1)} + \|B\|_{(\hat k_2,1)} \\
&\geq& \|X^\star\|_{(\hat k_1,1)} + \|B\|_{(\hat k_2,1)}\\
&:=& \left\|\begin{pmatrix} X^\star&0\\0 &B\end{pmatrix}\right\|_{(k,1)},
\end{eqnarray*}
where $k_1+k_2 = \hat k_1+\hat k_2 = k$ are suitable integers to fulfill 
the two definitions. % operators. 
Note that we have used the fact that the singular values of $\begin{pmatrix} X&0\\0 &B\end{pmatrix}$ are precisely the union of singular values of $X$ and $B$. Applying Lemma \ref{KyFan} once more completes the proof. \eproof

For an arbitrary matrix $B$ with rank $r$, we denote its thin SVD as
$B = U_B\Sigma_BV_{B}^*$. 
Define two projections $P_{B, \mathcal{L}} := U_{B} U_{B}^*$ 
and $P_{B, \mathcal{R}} := V_{B} V_{B}^*$. 
Let $U_B^\perp$ and $V_B^\perp$ be the orthogonal complements of 
$U_B$ and $V_B$, respectively.
\addtocounter{thm}{1}

\textbf{Simultaneous Block Assumption (SB):} Assume $(U_B^\perp)^* A V_C = 0$ and $U_B^* A V_C^\perp = 0$.
\begin{theoremappx} 
\label{tLEYappx}
Fix $\mathbb{N}\ni k\leq\rank(A)$. 
Under the SB assumption, $B^{\dag} (P_{B, \mathcal{L}} AP_{C, \mathcal{R}})_{(k)} C^{\dag}$ is the minimum Frobenius norm solution of 
\begin{equation}
\label{eLEYappx}
\min_{X: ~\rank{X}\leq k} \aui{A-BXC}.
\end{equation}
%It is the unique solution 
The solution is unique
iff the $k$-th and $(k+1)$-th largest singular values of $P_{B, \mathcal{L}} AP_{C, \mathcal{R}}$ differ.
\end{theoremappx}
\textbf{Proof:} 
Due to the SB assumption and the unitary invariance of the norm:
%we have %that
$$\aui{A-BXC} = \auibig{\begin{pmatrix} \hat A -\Sigma_B\hat X\Sigma_C & 0 \\ 0 & (U_B^\perp)^*AV_C^\perp \end{pmatrix}},$$ 
where $\hat A = U_B^*AV_C, \hat X = V_B^*XU_C$. 
It is apparent that $\rank(\hat X) \leq \rank(X) \leq k$. 

%By appealing to Lemma \ref{KyFan}, we can reduce the problem $\min\limits_{\rank{Z}\leq k} \ui{\begin{pmatrix} W - Z \\ 0 \end{pmatrix} }$ into $\min\limits_{\rank{Z}\leq k} \ui{W - Z}$. Indeed, suppose $Z^\star$ is the optimal solution for the latter under all unitarily invariant norms, then Lemma \ref{KyFan} ensures that $\forall k, \|W - Z^\star\|_{(k,1)} \leq \| W - Z\|_{(k,1)}$ for all feasible $Z: \rank(Z)\leq k$. However, we know that $\|\begin{pmatrix} Z \\ 0 \end{pmatrix}\|_{(k,1)} = \|Z\|_{(k,1)}$ since the two matrices have exactly the same singular values. Applying Lemma \ref{KyFan} again we have $\ui{\begin{pmatrix} W - Z^\star \\ 0 \end{pmatrix} } \leq \ui{\begin{pmatrix} W - Z \\ 0 \end{pmatrix} }$, for all feasible $Z: \rank(Z)\leq k$.

Next,
by Proposition \ref{propappx}, we need only consider $\min\limits_{\rank(X)\leq k} \aui{\hat A-\Sigma_B\hat X\Sigma_C}$. 
%Just apply Theorem \ref{tEYappx}: 
%Now 
Applying Theorem \ref{tEYappx} we obtain
$\Sigma_B\hat X\Sigma_C = (\hat A)_{(k)}$. 
Since $\Sigma_B$ and $\Sigma_C$ are invertible, one can easily recover 
$X = V_B \Sigma_B^{-1}(\hat A)_{(k)}\Sigma_C^{-1}U_C^*$ whose Frobenius norm 
is minimal (\cite{Penrose56}). 
It is straightforward to verify that our choice of $X$ indeed simplifies 
to the form given in the theorem. 
The uniqueness property is inherited from Theorem~\ref{tEYappx}. \eproof

\textbf{Simultaneous Diagonal Assumption (SD):} In addition to the SB assumption, assume furthermore $U_B^*AV_C$ is diagonal.%
\footnote{
We call  rectangular matrix $A$ diagonal if $A_{ij}\! =\! 0\; \forall i\!\ne\! j$.
}
\begin{theoremappx}
\label{tNEYappx}
Let $\lambda > 0$. Under the SD assumption, the matrix problem
\begin{equation}
\label{eNEYappx}
\min_{X} \ui{A-BXC} + \lambda\cdot\uip{X} 
\end{equation}
has an optimal solution of the form $X^\star = V_B\Sigma_X U_C^*$, with $\Sigma_X$ being diagonal.
\end{theoremappx}
\textbf{Proof:} Due to the SD assumption and the unitary invariance of the norm: $$\ui{A-BXC} = \uibig{\begin{pmatrix} \hat A-\Sigma_B\hat X\Sigma_C & 0 \\ 0 & (U_B^\perp)^*AV_C^\perp \end{pmatrix}}, $$ 
where $\hat A = U_B^*AV_C$ and $\hat X = V_B^*XU_C$. 
Fix $X$ and define $\tilde X := V_B Y U_C^*$, 
where $Y$ is obtained by zeroing out all components of $\hat X$ except 
the diagonal. 
We now argue that $\tilde X$ has smaller objective value than $X$.

Due to unitary invariance: 
\begin{eqnarray*}
\uip{\tilde X} 
& = & \uipbig{\begin{pmatrix} Y & 0 \\ 0 & 0\end{pmatrix} } 
\\
& \leq & \uipbig{\begin{pmatrix} V_B^*XU_C & 0 \\ 0 & 0\end{pmatrix}} 
\\
& \leq & \uip{X}, 
\end{eqnarray*}
where the inequalities follow from Lemma \ref{diagdom}. 
Since $\hat A$ is assumed %to be 
diagonal, one can use similar arguments as in Proposition \ref{propappx} to show that $\ui{A-B\tilde XC} \leq \ui{A-BXC}$. 

Therefore we may restrict our attention to matrices in the form of $\tilde X:= V_B Y U_C^*$, where $Y$ is everywhere zero except on its diagonal. 
But then (\ref{eNEYappx}) reduces to 
$$\!\!\!\!\!\min_{y} \uibig{\begin{pmatrix} \hat A-\Sigma_B\diag(y)\Sigma_C & 0 \\ 0 &0 \end{pmatrix}} + \lambda\cdot\uipbig{\begin{pmatrix} \diag(y) &0 \\ \!0 &0 \end{pmatrix}},$$
which is a vector problem.
\eproof

\begin{theoremappx}
\label{tranknormappx}
Let $\lambda > 0$. 
Then $\exists~ r\in\{0,\ldots,\rank(\hat A)\}$ such that $V_B(\hat A - \hat A_{(r)})U_C^*$ is the minimum Frobenius norm solution of
\begin{equation}
\label{eranknormappx}
\min_X \rank(BAC-BXC) + \lambda \rui{X},
\end{equation}
where $\rui{\cdot}$ is either the rank function or a unitarily invariant norm.
\end{theoremappx}
\textbf{Proof:} 
Let us first consider $\rui{\cdot} = \ui{\cdot}$. 

\emph{Step 1:} Due to unitary and scaling invariance and Lemma~\ref{diagdom},
we have: 
\begin{eqnarray*}
\rank[B(A-X)C] &=& \rank(\hat A - \hat X), \\
\lambda \ui{X}  &\geq& \lambda \uip{\hat X},
\end{eqnarray*}
where $\hat X = V_B^*XU_C$, and $\uip{\hat X} := \uibig{\begin{pmatrix} \hat X & 0 \\ 0 & 0\end{pmatrix}}$ is easily verified to be a unitarily invariant norm. Therefore we need only consider  $$\min_{\hat X} \rank(\hat A - \hat X) + \lambda \uip{\hat X}.$$ 

\emph{Step 2:} We now argue that we may restrict $\hat X$ to have the same singular matrices as $\hat A$. Introduce $Z = \hat A - \hat X$ and consider
$$\min_Z \rank(Z) + \lambda\uip{\hat A - Z}.$$
As indicated in Remark 2 in the main body of the paper, 
this rank regularized problem can be solved by considering a sequence of rank constrained problems. But, by Theorem~\ref{tEYappx}, the optimal solution of each rank constrained problem can be chosen to have the same singular matrices as $\hat A$. Therefore the optimal $Z$, hence $\hat X$, can be so chosen as well.

\emph{Step 3:} To determine the singular values of $\hat X$, we observe that unitarily invariant norms are always increasing functions of the singular values \citep{Bhatia97}. Given the value of $\rank(\hat A -\hat X)$, say $r$, then $\tilde X := \hat A -\hat A_{(r)}$ is easily seen to be optimal. But $r$ can only take a few integral values. 

\emph{Step 4:} Finally, given $\hat X$, we may easily recover $X = V_B \hat X U_C^*$ which is guaranteed to have minimum Frobenius norm (\cite{Penrose56}). The proof for $\rui{\cdot} = \ui{\cdot}$ is complete.

Now consider $\rui{\cdot} = \rank(\cdot)$. Step 1 clearly remains true, hence we need only consider 
$$\min_{\hat X} \rank(\hat A - \hat X) + \lambda\cdot \rank(\hat X).$$
Let $\hat X^\star$ be a minimizer with rank $(\rank(\hat A) - r)$, then we see that $\tilde X := \hat A - \hat A_{(r)}$ must also be optimal since 
$$\rank(\tilde X) = (\rank(\hat A) - r) = \rank(\hat X^\star),$$
$$\rank(\hat A - \tilde X) = r = \rank(\hat A) - \rank(\hat X^\star) \leq \rank(\hat A - \hat X^\star).$$
From the optimality of $\hat X^\star$ we also conclude that $\rank(\hat A - \hat X^\star) = r$. 
But then $\tilde X$ must have smaller Frobenius norm than $\hat X^\star$ since the former is an optimal solution of 
$$\min_{Y:~ \rank(\hat A - Y) = r} \fro{Y},$$
while the latter is a feasible solution.
\eproof

\end{document}